\documentclass[sigconf]{acmart}

\usepackage{booktabs} 
\usepackage{subcaption}

\setcopyright{rightsretained}

\acmDOI{10.475/123_4}

\acmISBN{123-4567-24-567/08/06}

\acmArticle{4}
\acmPrice{15.00}

\begin{document}
\title{DeepNorm - A Deep learning approach to Text Normalization}


\author{Shaurya Rohatgi}
\affiliation{%
  \institution{Pennsylvania State University}
  \streetaddress{University Park}
  \city{State College} 
  \state{Pennsylvania} 
  \postcode{16801}
}
\email{szr207@ist.psu.edu}

\author{Maryam Zare}
\affiliation{%
  \institution{Pennsylvania State University}
  \streetaddress{University Park}
  \city{State College} 
  \state{Pennsylvania} 
  \postcode{16801}
}
\email{muz50@psu.edu}


\begin{abstract}
This paper presents an simple yet sophisticated approach to the challenge by Sproat and Jaitly (2016) 
- given a large corpus of written text aligned to its normalized spoken form, train an
RNN to learn the correct normalization function. Text normalization for a token seems very straightforward without it's context. But given the context of the used token and then normalizing becomes tricky for some classes. We present a novel approach in which the prediction of our classification algorithm is used by our sequence to sequence model to predict the normalized text of the input token. Our approach takes very less time to learn and perform well unlike what has been reported by Google (5 days on their GPU cluster). We have achieved an accuracy of 97.62 which is impressive given the resources we use. Our approach is using the best of both worlds, gradient boosting - state of the art in most classification tasks and sequence to sequence learning - state of the art in machine translation. We present our experiments and report results with various parameter settings. 
\end{abstract}

%
%



\maketitle

\section{Introduction}
\begin{figure*}[h]
\centering
  \begin{subfigure}[b]{0.5\linewidth}
    \centering
    \includegraphics[width=\textwidth,scale=0.5]{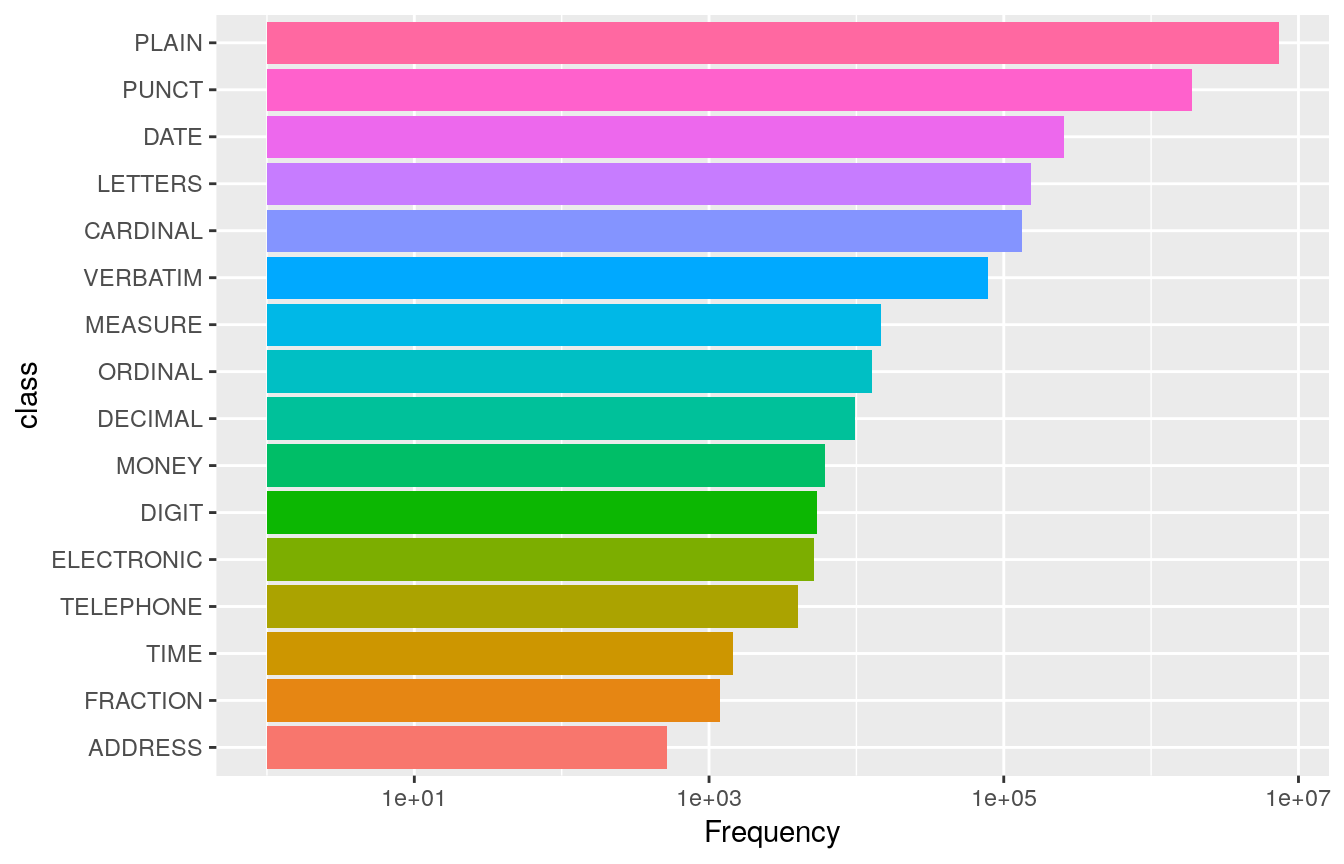}
    \caption{Semiotic Class Distribution}
  \end{subfigure}%
  \begin{subfigure}[b]{0.5\linewidth}
    \centering
    \includegraphics[width=\textwidth,,scale=0.5]{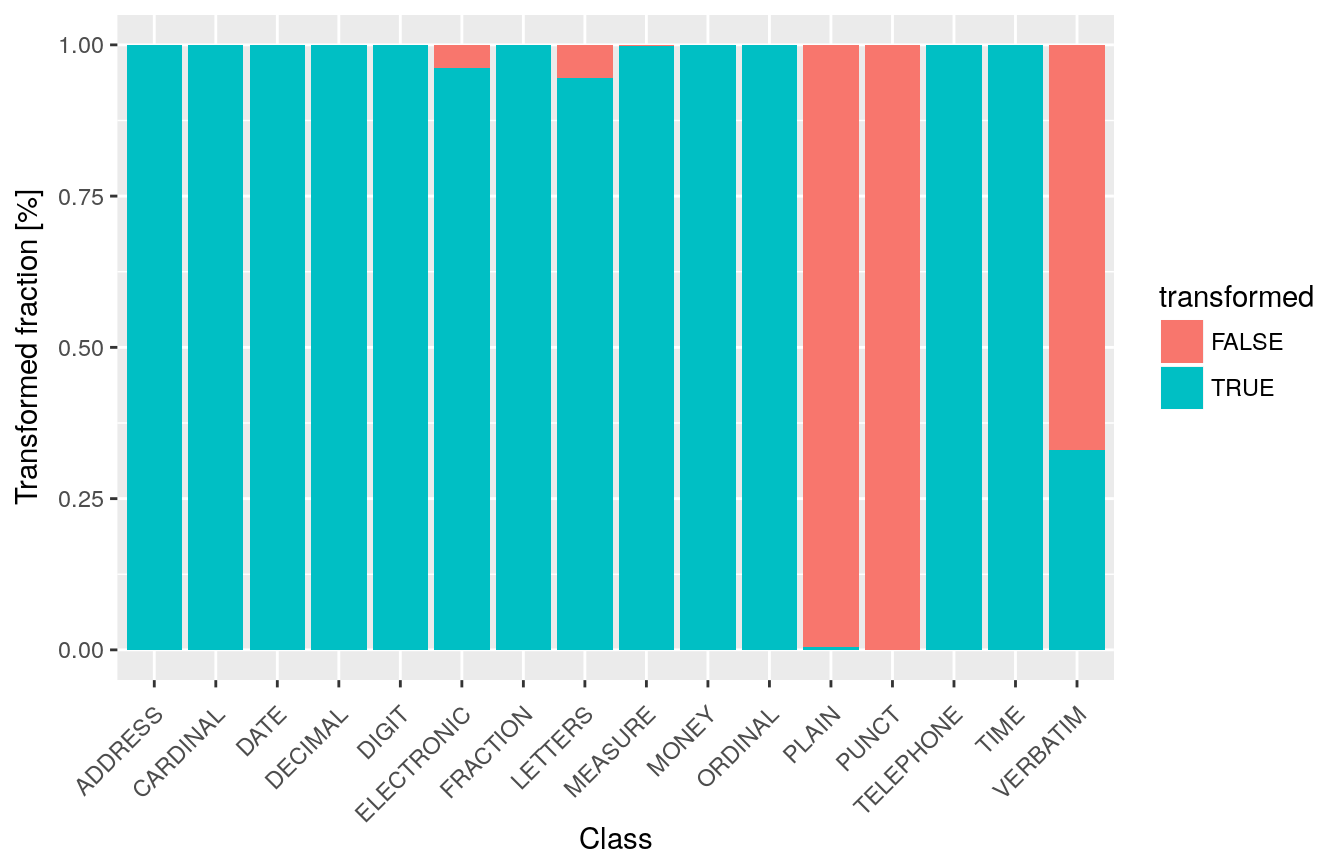}
    \caption{Tokens to be Transformed vs Non-Transformed}
  \end{subfigure}\\%
\caption{Train Data Semiotic Class Analysis - Source Kaggle}\label{fig:prob1b}
\end{figure*}
Within the last few years a major shift has taken
place in speech and language technology: the field
has been taken over by deep learning approaches. For
example, at a recent NAACL conference well more
than half the papers related in some way to word embeddings or deep or recurrent neural networks.
This change is surely justified by the impressive
performance gains to be had by deep learning, something that has been demonstrated in a range of areas from image processing, handwriting recognition, acoustic modeling in automatic speech recognition (ASR), parametric speech synthesis for text-to-speech (TTS), machine translation, parsing, and go
playing to name but a few.
While various approaches have been taken and
some NN architectures have surely been carefully designed for the specific task, there is also a widespread
feeling that with deep enough architectures, and
enough data, one can simply feed the data to one's
NN and have it learn the necessary function.
In this paper we present an example of an application that is unlikely to be amenable to such a "turn-
the-crank" approach. The example is text normalization, specifically in the sense of a system that converts from a written representation of a text into a
representation of how that text is to be read aloud.
The target applications are TTS and ASR - in the
latter case mostly for generating language modeling
data from raw written text. This problem, while often
considered mundane, is in fact very important, and a
major source of degradation of perceived quality in
TTS systems in particular can be traced to problems
with text normalization.
\\ 
We start by describing the prior work in this area, which includes use of RNNs in text normalization. We describe the dataset provided by Google and Kaggle and then we discuss our approach and experiments \footnote{https://github.com/shauryr/google\_text\_normalization} we performed with different Neural Network architectures.
\section{Related Work}
Text normalization has a long history in speech technology, dating back to the earliest work on full TTS
synthesis (Allen et al., 1987). Sproat (1996)   provided
a unifying model for most text normalization problems in terms of weighted finite-state transducers
(WFSTs). The first work to treat the problem of text
normalization as essentially a language modeling
problem was (Sproat et al., 2001 ) . More recent machine learning work specifically addressed to TTS
text normalization include (Sproat, 2010; Roark and
Sproat, 2014; Sproat and Hall, 2014).

In the last few years there has been a lot of work
that focuses on social media (Xia et al., 2006; Choudhury et al., 2007; Kobus et al., 2008; Beaufort et
al., 2010; Kaufmann, 2010; Liu et al., 2011; Pennell and Liu, 2011; Aw and Lee, 2012; Liu et al.,
2012a; Liu et al., 2012b; Hassan and Menezes, 2013;
Yang and Eisenstein, 2013). This work tends to focus on different problems from those of TTS: on
the one hand one, in social media one often has to
deal with odd spellings of words such as "cu 18r", "coooooooooooooooolllll", or "dat suxx", which are
less of an issue in most applications of TTS; on the
other, expansion of digit sequences into words is critical for TTS text normalization, but of no interest to
the normalization of social media texts.

Some previous work, also on social media normalization, that has made use of neural techniques includes (Chrupała, 2014; Min and Mott, 2015). The
latter work, for example, achieved second place in
the constrained track of the ACL 2015 W-NUT Normalization of Noisy Text (Baldwin et al., 2015),
achieving an F1 score of 81.75\%.

\section{Dataset}
The original work by Sproat and Jaitly 
uses 1.1 billion words for English text and 290 words for Russian text. In this work we used a subset of the dataset submitted by the authors for the Kaggle competition \footnote{https://www.kaggle.com/c/text-normalization-challenge-english-language} (table \ref{table:1}).
The dataset is derived from Wikipedia regions which could be decoded as UTF8. The text is then divided into sentences and through the Google TTS system's Kestrel text normalization system to produce the normalized version of that text. A snippet is shown in the figure 1 
. As described in (Ebden and Sproat, 2014),
Kestrel's verbalizations are produced by first tokenizing the input and classifying the tokens, and
then verbalizing each token according to its semiotic class. The majority of the rules are hand-built using the Thrax finite-state grammar development system (Roark et al., 2012). Most ordinary
words are of course left alone (represented here as
\textit{<self>}), and punctuation symbols are mostly transduced to \textit{<sil>} (for "silence").

Sproat and Jaitly 
report that a manual analysis of about 1,000 examples from
the test data suggests an overall error rate of approximately 0.1\% for English. Note that although the test data were of course
taken from a different portion of the Wikipedia text
than the training and development data, nonetheless
a huge percentage of the individual tokens of the test data 99.5\% in
the case of English - were found in the training set.
This in itself is perhaps not so surprising but it does
raise the concern that the RNN models may in fact be
memorizing their results, without doing much generalization.
\begin{table}[h!]
\centering
 \begin{tabular}{||c  c||} 
 \hline
 Data & No. of tokens \\ [0.5ex] 
 \hline
 Train  & 9,918,442 \\ 
 \hline
 Test  & 1,088,565 \\ [1ex]  
 \hline
\end{tabular}
\caption{Kaggle Dataset}
\label{table:1}
\end{table}

\subsection{Data Exploratory Analysis}
\begin{figure*}
  \includegraphics[scale=0.5]{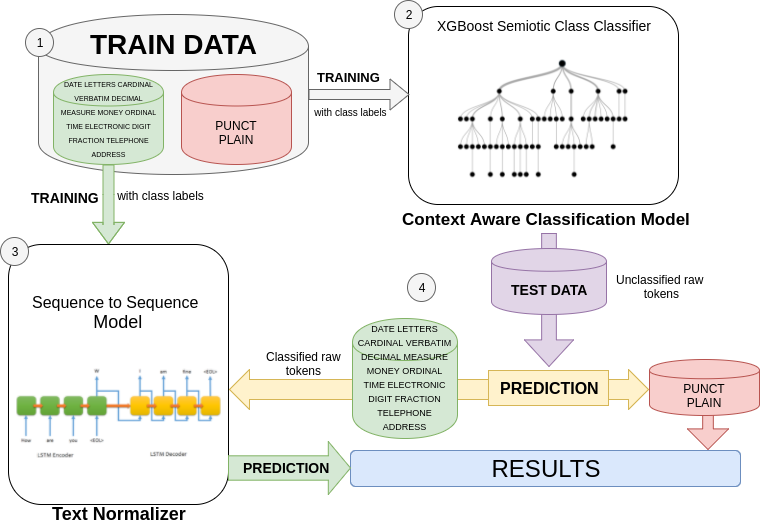}
  \caption{Our Model for Kaggle's Text Normalization Challenge}\label{fig:train}
\end{figure*}

In total, only about 7\% of tokens in the training data, or about $660k$ objects in total, were changed during the process of text normalization in the train data. This explains the high baseline accuracies we can achieve even without any adjustment of the test data input.

The authors of the challenge refer the classes of tokens as semiotic classes. The classes can be seen in the Figure \ref{fig:prob1b}. In total there are 16 classes. The "PLAIN" class is by far the most frequent, followed by "PUNCT" and "DATE". "TIME", "FRACTION", and "ADDRESS" having the lowest number of occurrences (around/below 100 tokens each).

Over exploring the dataset we find that "PLAIN" and "PUNCT" semiotic classes do not need transformation or they need not be normalized. We exploit this fact to our advantage when we train our sequence to sequence text normalizer by only feeding the tokens which need normalization. This reduces the burden over our model and filters out what may be noise for our model. This is not to say that notable fraction of "PLAIN" class text elements did change. But the fraction was too less to be considered for training our model. For example, "mr" to "mister" or "No." to "number".
We also analyzed the length of the tokens to be normalized in the dataset. We find that short strings are dominant in our data but longer ones with up to a few 100 characters can occur. This was common with "ELECTRONIC" class as it contains URL which can be long.

\begin{figure*}
  \includegraphics[scale=0.4]{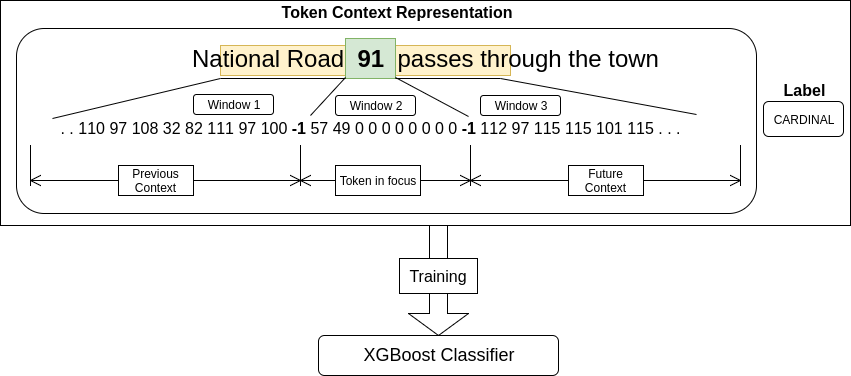}
  \caption{Context Aware Classification Model - XGBoost Semiotic Class Classifier
}\label{fig:classifier}
\end{figure*}

\section{Baseline}
As mentioned above, most of the tokens in the test data are similar to those in the test data. We exploited this fact to hold the data in the train set in memory and predicted the class of the token using the train set. 

We have written a set of 16 functions for every semiotic class to normalize it. Using the predicted class we used the regular expression functions  to normalize the test data. We understand this is not the correct way to do this, but it provides a very good and competitive baseline for our algorithm. We score 98.52\% on the test data using this approach. This also defines a line whether our model is better or worse than memorizing the data.

\section{Methodology}
Our approach involves modeling the problem as classification and translation problem. The model has two major parts, a classifier which determines the tokens that need to be normalized and a sequence to sequence model that normalizes the non standard tokens (Figure \ref{fig:train}). We first explain training and testing process, then we explain classifier and sequence to sequence models in more detail.\\
Figure ~\ref{fig:train} shows the whole process of training and testing. We trained classifier and sequence to sequence model individually and in parallel. Training set has 16 classes, 2 of which don't need any normalization, so we separated tokens from those two classes from others and only fed tokens from remaining 14 classes to the sequence to sequence model. On the other hand classifier is trained on the whole data set since it need to distinguish between standard and non standard tokens.\\
Once training is done, we have a two stages pipeline ready. Raw data is fed to the classifier. Results of classifier are two sets of tokens. Those that don't need to be normalized are left alone. Those that need to be normalized are passed to the sequence to sequence model. Sequence to sequence model converts the non standard tokens to standard forms. Finally the output is merged with tokens from the classifier that were marked as standard ones as the final result.\\
Now we explain both classifier and normalizer in more detail.

\subsection{Context Aware Classification Model (CAC) }
Detecting the semiotic class of the token is the key part of this task. Once we have determined the class of a token correctly, we can normalize the it accordingly.
The usage of a token in a sentence determines its semiotic class. To determine the class of the token in focus, the surrounding tokens play an important role. Specially in differentiating between classes like DATE and CARDINAL, for example, CARDINAL 2016 is normalized as two thousand and sixteen, while DATE 2016 is twenty sixteen, the surrounding context is very important.

Our context aware classification model is explained in the Figure \ref{fig:classifier}
We choose a window size $k$ and we represent every character in the token with it's ASCII value. We pad the empty window with zeros. We use the preceding $k$ characters of the tokens and the later $k$ characters of the tokens around the token in focus. This helps the classifier understand in which context the token in focus has been used. 
We use vanilla gradient boosting algorithm without any parameter tuning. Other experiment details are in the next section.


\begin{table}
\centering
\begin{tabular}{|c|c|}
\hline
\textbf{Window Size} & \textbf{Dev Set Accuracy}   \\ \hline
10 & \textbf{99.8087}  \\ \hline
20 & 99.7999   \\ \hline
40 & 99.7841   \\ \hline
\end{tabular}
\caption{Context aware classification model - Varying Window size}
\label{table:window}
\end{table}

\subsection{Sequence to Sequence Model}
In this section we explain the sequence to sequence model in detail. We used a 2-layer LSTM reader that reads input tokens, a layer of 256 attentional units, an embedding layer, and a 2-layer decoder that produces word sequences. We used Gradient Descent with decay as an optimizer.\\ 
The encoder gets the input  $(x_1,x_2,...,x_{t1})$ and decoder gets the inputs encoded sequence $(h_1,h_2,...,h_{t1})$ as well as the previous hidden state $s_{t-1}$ and token $y_{t-1}$ and outputs $(y_1,y_2,...,y_{t2})$. The following steps are executed by decoder to predict the next token:
\begin{equation}
\begin{aligned}
r_t = \sigma(W_ry_{t-1}+U_rs_{t-1}+C_rc_t)\\
z_t = \sigma(W_zy_{t-1}+U_zs_{t-1}+C_zc_t)\\
g_t = tanh(W_py_{t-1}+U_p(r_t \circ s_{t-1})+C_pc_t)\\
s_t = (1-z_t) \circ s_{t-1} +z_t \circ g_t \\
y_t = \sigma(W_oy_{t-1}+U_o s_{t-1}+C_oc_t)\\
\end{aligned}
\end{equation}

\begin{figure}
  \includegraphics[scale=0.4]{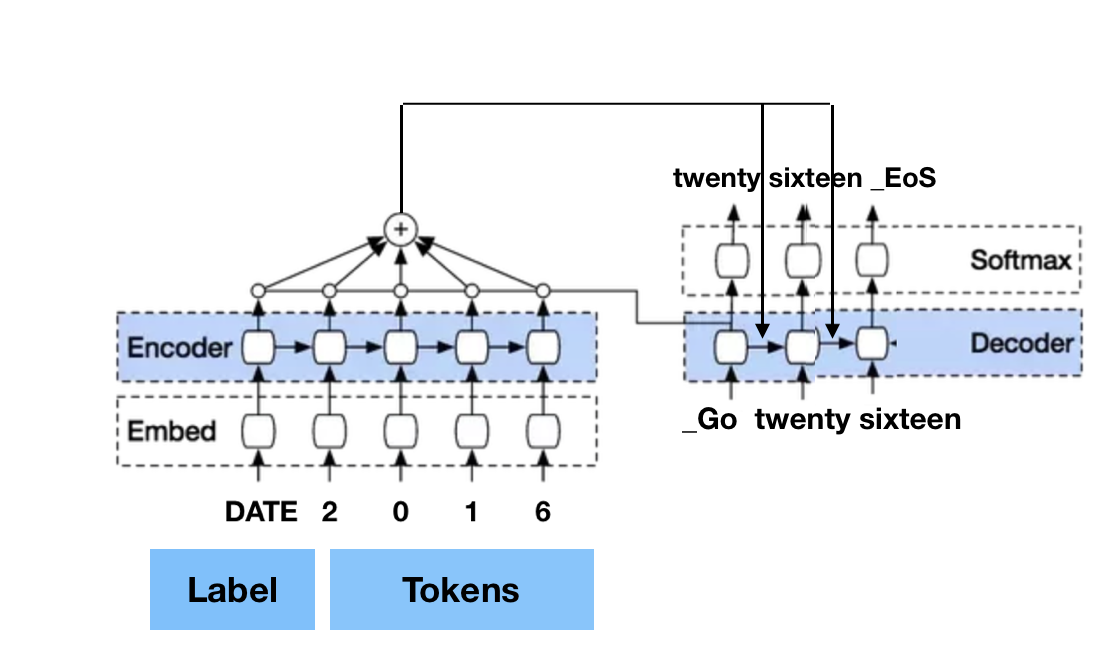}
  \caption{Sequence to Sequence Model}\label{fig:seq2seq}
\end{figure}
The model first computes a fixed dimensional representation context vector $c_t$ , which is the weighted sum of the encoded sequence. Reset gate, r, controls how much information from the previous hidden state $s_{t-1}$ is used to create a proposal hidden state. The update gate, z, controls how we much of the proposal we use in the new hidden state $s_t$.  Finally we calculate the t-th token using a simple one layer neural network using the context, hidden state, and previous token. \\
We fed tokens in a window of size of 20 with the first one being the label (Figure \ref{fig:seq2seq}). For example, if we want to get the normalized form of 2017 we will feed it in the following form <label> <2> <0> <1> <7> <PAD> ... <PAD>. In cases where the input size is less than 20 we fill the empty spots with reserved token,  <PAD>. Batch size is set to 64 and the vocabulary size is 100,000. We tried smaller vocabulary sizes but since our data set is very sparse we didn't get a good accuracy, after making it bigger the accuracy improved significantly. \\
\begin{table}[]
\centering
\begin{tabular}{|c|c|c|}
\hline
           & \textbf{Google's RNN}                 & \textbf{CAC+Seq2seq}                  \\ \hline
All        & 0.995                                 & 0.9762                                \\
PLAIN      & 0.999                                 & -                                     \\
PUNCT      & 1.00                                  & -                                     \\
DATE       & 1.00                                  & \textbf{0.998}                        \\
LETTERS    & 0.964                                 & 0.818                                 \\
CARDINAL   & 0.998                                 & \textbf{0.996}                        \\
VERBATIM   & 0.990                                 & {\color[HTML]{FE0000} \textbf{0.252}} \\
MEASURE    & {\color[HTML]{FE0000} \textbf{0.979}} & 0.955                                 \\
ORDINAL    & 1.00                                  & 0.982                                 \\
DECIMAL    & 0.995                                 & \textbf{0.993}                        \\
ELECTRONIC & 1.00                                  & {\color[HTML]{FE0000} \textbf{0.133}} \\
DIGIT      & 1.00                                  & \textbf{0.995}                        \\
MONEY      & {\color[HTML]{FE0000} \textbf{0.955}} & 0.824                                 \\
FRACTION   & 1.00                                  & 0.847                                 \\
TIME       & 1.00                                  & 0.872                                 \\
ADDRESS    & 1.00                                  & 0.931                                 \\ \hline
\end{tabular}

\caption{Classwise Accuracy comparison with Google's RNN - Our model comes close in some classes to the existing state of the art deep learning model}
\label{tab:compres}
\end{table}
\section{Experiments and Results}
\subsection{Classification}
For classification we use random forests, with the default parameters and early stopping. We used XGBoost \footnote{http://xgboost.readthedocs.io/en/latest/get\_started/index.html} module for python. Table \ref{table:window}
shows the results for different window sizes. We used a 10\% of the train data as our validation set. Training this a classifier on 9 million tokens takes a lot of time to train, of the order of 22 hours.

We see an interesting behavior, as the window size is decreased the classifier's accuracy also increases. This behavior is reasonable as most of the tokens are short (less than 10) in length. Also, the starting characters and the surrounding context of the long tokens are enough to determine their semiotic class.
Once we have trained this classifier we predict the classes for the test data and label each token with it's semiotic class. These labeled tokens are then normalized by the sequence to sequence model, which we discuss in the following section.
\begin{table}
\centering

\begin{tabular}{|c|c|c|}
\hline
                   & \textit{2 Layers} & \textit{3 Layers} \\ \hline
\textit{64 Nodes}  & 97.46             & 97.41             \\ \hline
\textit{128 Nodes} & 97.55             & 97.53             \\ \hline
\textit{256 Nodes} & \textbf{97.62}    & 97.6              \\ \hline
\end{tabular}
\caption{Accuracy on Test data - Experiments with varying number of nodes and layers}
\label{tab:res}
\end{table}
\subsection{Sequence to Sequence Model}
We build our model using tensorflow's \footnote{https://www.tensorflow.org/} python module. Here are the other details about the model -
\begin{itemize}
\item Number of encoder/decoder layers: 2-3
\item One layer of embedding layer
\item Number of hidden units: 256-128-64
\item Encoder size: 20
\item Decode size: 25
\item latent space representation size : 256
\item Vocabulary size: 100,000
\item Optimizer: Gradient Descent with learning decay
\item Number of Epochs: 10
\end{itemize}
\begin{table*}
\centering
\begin{tabular}{ccc}
\hline
\multicolumn{1}{|c|}{\textbf{Semiotic Class}} & \multicolumn{1}{c|}{\textbf{before}} & \multicolumn{1}{c|}{\textbf{after (predicted)}} \\ \hline
\multicolumn{1}{|c|}{DATE} & \multicolumn{1}{c|}{2016} & \multicolumn{1}{c|}{twenty sixteen} \\ \hline
\multicolumn{1}{|c|}{CARDINAL} & \multicolumn{1}{c|}{2016} & \multicolumn{1}{c|}{two thousand and sixteen} \\ \hline
\multicolumn{1}{|c|}{DIGIT} & \multicolumn{1}{c|}{2016} & \multicolumn{1}{c|}{two o one six} \\ \hline
 &  &  \\ \hline
\multicolumn{1}{|c|}{CARDINAL} & \multicolumn{1}{c|}{1341833} & \multicolumn{1}{c|}{one million three hundred fourteen thousand eight hundred thirty three} \\ \hline
\multicolumn{1}{|c|}{TELEPHONE} & \multicolumn{1}{c|}{0-89879-762-4} & \multicolumn{1}{c|}{o sil eight nine eight seven seven sil nine six two sil four} \\ \hline
\multicolumn{1}{|c|}{MONEY} & \multicolumn{1}{c|}{14 trillion won} & \multicolumn{1}{c|}{fourteen won} \\ \hline
 &  &  \\ \hline
\multicolumn{1}{|c|}{VERBATIM} & \multicolumn{1}{c|}{$\omega$} & \multicolumn{1}{c|}{\textit{w m s b}} \\ \hline
\multicolumn{1}{|c|}{LETTERS} & \multicolumn{1}{c|}{mdns} & \multicolumn{1}{c|}{\textit{c f t t}} \\ \hline
\multicolumn{1}{|c|}{ELECTRONIC} & \multicolumn{1}{c|}{www.sports-reference.com} & \multicolumn{1}{c|}{\textit{w w r w dot t i s h i s h e n e n e dot c o m}} \\ \hline
\multicolumn{1}{l}{} & \multicolumn{1}{l}{} & \multicolumn{1}{l}{}
\end{tabular}
\caption{Results Analysis of Seq2Seq Model - The prediction gets worse as we go down the table. }
\label{table:analysis_seq2seq}
\end{table*}

Every other parameter used was default parameter provided by tensorflow framework. \\
Table \ref{tab:res} shows the accuracy on test data increases significantly as we increase the number of nodes on the encoder side. We also see that increasing the number of layers has very little effect.
We wanted to experiment with more nodes but given the time and resources we could only experiment with these parameter settings. The test data had approximately 60,000 tokens (needed to be normalized), and using such a model to predict the normalized version of the test tokens took about 6 hours. 
We present the class-wise comparison of the results in the table \ref{tab:compres}. One thing to note here is that we evaluated our data on a 600,000 samples but Google does it only for 20,000 samples. We can see that our model performs nearly as well as Google's RNN. But our model also suffers in classes such as VERBATIM and ELECTRONIC. As discussed below, VERBATIM has special characters from different languages and we chose only the top 100,000 in our vocabulary (GPU memory constraints). We think that if the vocabulary size is increased we can achieve far better results. Also for ELECTRONIC class the window size of the encoder input was the constraint. We can see from table \ref{table:analysis_seq2seq} that it starts well but as the sequence gets longer it predicts irrelevant characters. We believe increasing the encoder sequence length can improve this aspect of our model. \\
Table \ref{table:analysis_seq2seq} shows the results. For three classes DATE, CARDINAL, and DIGIT the model works very well, and the accuracy is very close to Google's model. For example in case of token '2016', it is shown that the model can distinguish different concepts very well and outputs the correct tokens. We think this is because we are feeding the label with the tokens to the sequence to sequence model, so it learns the differences between these classes pretty good.\\
The next three classes are showing acceptable results. Model shows some difficulties in telephone numbers, big cardinal numbers, and class MONEY. Errors are not very bad. In most cases usually one word is missed or the order is reversed.\\
We got low accuracy on the last three classes shown in Table \ref{table:analysis_seq2seq}. We see that Verbatim and Electronic classes have the lowest accuracy. For Verbatim we think the reason is the size of vocabulary. Since this class consists of special characters that have low frequency in the data set, a larger vocabulary could have improve the accuracy a lot. For Electronic class we think a larger encoder size can be very helpful. This class has tokens of up to length 40, which don't fit to the encoder we used.

\section{Conclusion}
In this project we proposed a model for the task of normalization.We present a context aware classification model and how we used it to clear out "noisy" samples. We then discuss our unique model, which at it's core is a sequence to sequence model which takes in the label and the input sequence and predicts the normalized sequence based on the label. We share our insights and analysis with examples of where our models shines and where we can improve. We also list out possible ways of improving the results further. We compare our results with the state of the art results and show that given limited computation power we can achieve promising results This project helped us understand sequence to sequence models and the related classification tasks very well. We also learned how much parameter tuning can effect the results and small changes makes big difference. We can also try Bidirectional RNNs as we saw if the sequence was longer the model was not accurate. 

Finally, we conclude that higher accuracy can be achieved via having a very good classifier. Classifier has an important role in this model and there is still lots of room for improvement. Using LSTM instead of XGBoost could have make the classifier stronger. But we rested our focus mostly on the sequence to sequence model as we wanted to understand and implement it. Due to the lack of time and limited resources we couldn't try this and we list this as a future work.


\bibliography{bib_file} 

\end{document}